# 基于BERT的非结构化领域文本知识抽取

王梓嘉[1] 李烨[2] 朱忠凯[2]

（1.中国科学院大学，北京 100000；2.微软亚太研发集团商用人工智能部，北京 100080）

**摘要：**随着知识图谱技术的发展和商业应用的普及，从各类非结构化领域文本中提取出知识图谱实体及关系数据的需求日益增加。这使得针对领域文本的自动化知识抽取颇有意义。本文提出了一种基于BERT的知识抽取方法，用于从非结构化的特定领域文本（例如保险行业的保险条款）中自动抽取知识点，以达到在构建知识图谱的过程中节约人力的目的。不同于常用的按照规则、模板或基于实体抽取模型的知识点抽取方法，本文将领域文本的知识点转化为问答对，以答案所在的位置前后的文本为上下文，以BERT基于SQuAD数据进行阅读理解任务的方式进行微调。用微调后的模型从更多保险条款中自动抽取知识点，获得了良好的效果。

# BERT-based knowledge extraction method of unstructured domain text

WANG Zijia[1], LI Ye[2], and ZHU Zhongkai[2]

(1. University of Chinese Academy of Sciences, Beijing, 100000, China; 2. Department of AI Vertical in Microsoft Software Technology Center Asia (STCA), Beijing, 100080, China)

**Abstract:** With the development and business adoption of knowledge graph, there is an increasing demand for extracting entities and relations of knowledge graphs from unstructured domain documents. This makes the automatic knowledge extraction for domain text quite meaningful. This paper proposes a knowledge extraction method based on BERT, which is used to extract knowledge points from unstructured specific domain texts (such as insurance clauses in the insurance industry) automatically to save manpower of knowledge graph construction. Different from the commonly used methods which are based on rules, templates or entity extraction models, this paper converts the domain knowledge points into question and answer pairs and uses the text around the answer in documents as the context. The method adopts a BERT-based model similar to BERT's SQuAD reading comprehension task. The model is fine-tuned. And it is used to directly extract knowledge points from more insurance clauses. According to the test results, the model performance is good.

## 0 引言

近年来随着各行业数字化转型的深入，相关电子文本数量与日激增。与此同时，越来越多的企业开始重视数据分析、挖掘以及数据资源的开发利用，诸如知识图谱，智能对话等计算机应用系统已经成为了各类企事业单位对内对外提供服务的基础。这类应用往往需要从各类非结构化领域文本中提取出其蕴含的结构化信息，用于数字化知识库的构建。

数据是计算机产品和服务的基础，为计算机提供数据成为了新时期企事业单位发展的新任务。企事业单位中原有的各类商务、业务文档资料蕴含了丰富的知识和信息，却都是为了人类阅读而撰写，相对于计算机程序的需求，多了很多的冗余信息。

目前，在应用这类数据时，基本上需要投入大量人力，通过阅读文档人工抽取出所需信息，并将其表示成计算机能够读取（“理解”）的形式。如此造成了许多额外的学习成本和人力资源消耗。

如何采用自动化的手段从非结构化的文本数据中发现知识，用以作为各种智能化应用所依存的数据资源，是知识抽取领域的研究热点。

本文以特定领域的非结构化文本为研究对象，提出了通过基于深度学习的语言理解模型对其进行知识抽取的方法。这一方法将待抽取知识点以问题-答案对形式呈现，并以人工标注数据作为训

练数据，在预训练模型的基础上进行迁移学习，通过微调（Fine Tuning）获得在同领域文本上自动抽取知识点的模型。

本文接下来的组织结构如下:第 1 节介绍国内外研究现状和相关工作，第 2 节介绍本文提出的基于深度学习的非机构文本知识抽取模型，第 3 节是实验验证和结果分析,最后是论文总结和未来工作。

## 1 相关工作

非结构化文本的知识抽取任务可以根据具体的知识结构以及应用场景从不同的角度进行算法设计，简单来说可以分为如下四类。

### 1.1 基于规则的知识抽取

对于具有统一结构规范的文档，可以采用构建规则的方式进行知识抽取。规则的构建往往通过人工的归纳总结来完成——即阅读大量同领域文本,从中选取,总结出最终的抽取规则。Jyothi[1]等人使用基于规则的方式从大量个人简历中抽取有效信息,构建数据库。JunJun[2]等人用类似的方法从学术文献中提取学术概念知识。这种方法的优点是不需训练模型,简单高效;缺点也很明显，我们构建的规则仅适用于相同结构的文本，而且必须具有较严格的格式规范，一旦文本结构稍有改变，就需要人工构建新的知识抽取规则，因此该方法不具备可移植性。

### 1.2 基于实体提取的知识抽取

知识抽取的一种任务称为实体提取，即从文本中抽取预先定义的标签内容,如时间,地点等,具体标签根据应用而定，其中最常用的一种知识抽取称为命名实体识别（named entity recognition, NER)。实体抽取本身可直接作为序列标注任务来解决，后者可使用传统的统计学习方法，如隐马尔可夫模型(HMM)[3],或条件随机场(CRF)[4]处理。

近年来一些深度学习模型也被应用到这类问题中,如结合了 BiLSTM 和 CRF 的序列标注方法曾取得很好的效果[5]。Lample[6]等人提出了一种新的网络结构，使用堆叠的 LSTM 来表示一个栈结构，直接构造多个单词的表示，并与 LSTM-CRF 模型做了比较。Ma[7]等人提出了基于 BiLSTM-CNN-CRF 的端到端序列标注模型。此外，微调后的 BERT 模型也能够在序列标注任务上达到很好的效果[8]。

### 1.3 基于 schema 的知识抽取

除了从文本中提取实体，实体之间的关系也是知识抽取的关注点，通常把实体及其关系组成三元组<E1, R, E2>，那么任务目标为从文本中抽取所有可能的实体关系三元组，其关系限定在预先设定的 schema 之内。

Zeng[9]等人设计了 CNN 来进行关系分类，但并非三元组。Makoto[10]等人通过构建基于 BiLSTM 和 Bi-TreeLSTM 的堆栈网络同时进行实体抽取和关系检测，从而实现实体关系的端到端预测。Li[11]等人采用一个编码器-解码器架构的双层 LSTM，构建了一个不局限于三元组形式的知识抽取模型，可以预测出固定格式的结构化知识。Zheng[12]等人通过一种标注策略把实体和关系提取任务转化为序列标注任务，而后构建与前面类似的 Bi-LSTM 模型来处理之。Luan[13]等人设计了一种多任务学习框架用于在科学文献中识别实体和关系，以构建科学知识图谱，此模型在没有任何领域先验知识的情况下优于现有模型。

### 1.4 基于问答的知识抽取

除了以上提到的知识抽取模式，一个不同的角度是将知识点本身看作一个问题，将知识点的内容作为该问题的答案，将知识点所在的文本段作为这个问答对的上下文，这样知识抽取模型便可以用问答模型来构造。

近年来，GPT[14]和 BERT[15]等预训练模型的出现使得这类问答阅读理解任务可以很好地作为其下游任务，仅需简单改造原有网络结构，并进行微调，即可得到很好的效果。Wang[16]等人在原始 BERT 的基础上使用多段落预测的方式改进了其在 SQuAD[17]数据集上的效果。Alberti[18]等人在 BERT 与 SQuAD 的基础上改进后，将其应用在一个更困难的问答数据集 NQ[19]上，$F_1$分数相对之前的基准线提升了 30%。

这种问答形式的知识抽取可以更灵活地处理不同结构的知识——只需将其定义为不同的问题，而不需要根据知识的形式单独设计新的网络结构。

## 2 基于深度学习的非结构化领域文本知识抽取

### 2.1 非结构化领域文本

不同行业的结构化文本因其行业特征而特色

迥异。有些行业的特定文档（例如医药说明书）不仅具备严格的结构而且在术语和用词上要求非常严格，比较适合基于规则的知识抽取。还有一些行业的文本与通用文本区别不大（例如新闻报道、访谈等），对其可以直接应用通用抽取技术。还有一些领域的文本，介于两者之间，有一定的专业性，但不很严格，不同企业的同类文本结构和措辞近似，但又有所差别，同一企业内部的术语使用和展示方式相对统一。保险行业的保险条款文档，就属于这第三类文本之列。

保险条款是保险合同双方当事人——保险人（保险公司）与投保人——共同约定的有关双方权利和义务的条文。一个保险条款一般由三部分组成：

1. 基本信息，即条款自身信息，包括：保险人，条款名称，条款简称，条款类型，期限类型，犹豫期，诉讼时效，备案号和备案时间，能否作为主险销售等；

2. 购买条件，即要本条款承接的被保险人需要具备的客观条件，包括：被保险人的年领、性别、职业/工种要求，体检要求，社保要求，必须如实告知的个人情况等；

3. 保险责任，也就是本条款的责任范围和赔付内容等；

对于同一个保险人而言，其制定的保险条款在各部分都会使用一致的术语和陈述。不同保险人之间的差异主要体现在保险责任，而基本信息和购买条件则往往很相似。如此的文本特征，为知识抽取提供了方便。

## 2.2 保险条款知识抽取方法选择

虽然保险条款具备一定程度的专业词汇，但专业词汇的使用大多没有业界统一标准（例如："犹豫期"又可称为"冷静期"等），而且条款文档作为交付给投保人阅读的文件，需要抽取的知识点大多混杂在一段自然语言表述之中，并不适合依据静态规则来进行文本抽取。

所需抽取知识点本身虽然可以采用实体抽取的方式来获取，但知识点对应的值却往往混杂在一段自然语言表述之中，无法与知识点描述一起被抽取出 。例如：某条款的诉讼时效是 2 年，这个"2 年"可能出现在下列这段描述中："受益人向我们请求给付保险金或保险费豁免的诉讼时效期间为 2 年，自其知道或应当知道保险事故发生之日起计算。"

因此在需要从保险条款中抽取基本信息、购买条件和保险责任等知识点时，我们就直接排除了基于规则和基于实体抽取的方法。

如果采用 Schema 式抽取，将知识点转化为三元组，则所需要的训练数据集和标注量都相对较大，相对于我们的目的而言，难免得不偿失。因此，我们最终选取了基于问答的知识抽取方式。

## 2.3 基于 BERT 的问答式知识抽取方式

近年来，基于预训练模型通过微调进行学习的方法在自然语言处理（Natural Language Process, NLP）领域取得了巨大成功，BERT 模型是其中的重要代表。

BERT 是一种基于转换器（transformer）的双向编码表示模型，它的拓扑结构是一个多层的双向转换器网络。BERT 模型是基于微调学习的典型应用，也就是说它的构建包含预训练和微调两个步骤。

首先在预训练阶段，对大量不同训练任务的未标记语料数据进行训练，将语料中的知识迁移进了预训练模型的文本嵌入（Embedding）中。

这样，在微调阶段，只需要在神经网络中增加一个额外的输出层，就可以对预训练模型进行调整了。具体而言，微调就是用预训练参数初始化 BERT 模型，然后，使用来自下游任务的标记数据对模型进行微调。

针对我们从保险文档中抽取知识点的需求，只需使用保险条款数据针对 BERT 的问答任务进行微调，就可以适应保险条款知识抽取的需求了。

## 2.4 基本方法

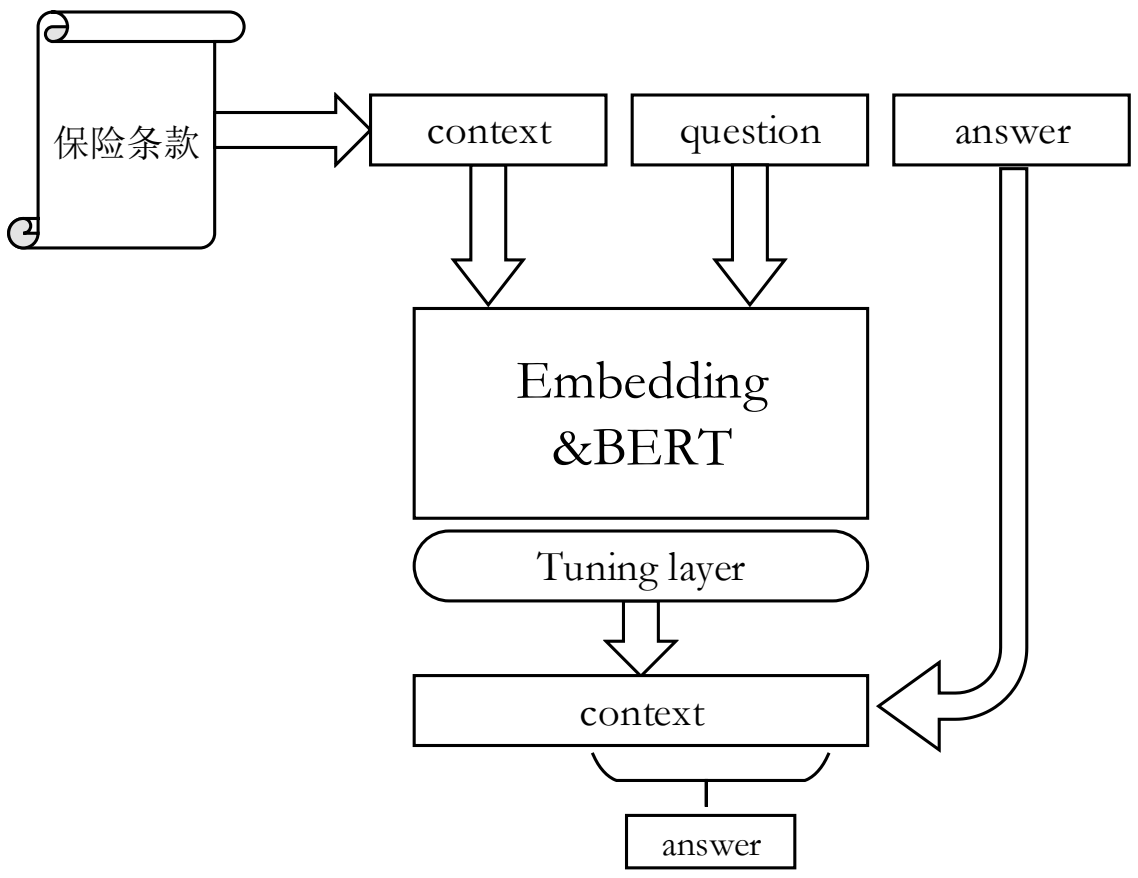

图 1 知识抽取模型

保险条款知识抽取过程首先将人工标注的保险条款知识点制作成<question, answer>的形式，然后使用文本解析程序将一份保险条款文档解析为一颗文档树，其中主标题为根节点，其后每一级标题都为上一级的子节点，而每段文字都被读

取为一个叶子节点。根据问题-答案对中的答案匹配到其所在叶子节点，以整个叶子节点对应的文本为该问题-答案对的上下文（context），最终构造成一个由<question, answer, context>组成的问答数据集，最后使用这一数据集对 BERT 预训练模型，按照针对 SQuAD 数据[20]进行阅读理解任务的微调方式进行训练，得到最终的知识抽取模型。如上图所示，对于问答任务，只需在 BERT 输出的编码向量后添加一层额外的全连接层，预测 answer 在 context 中的位置即可。

在测试时，对于新的保险条款，需要以同样的方式分析不同知识点问题所在的上下文，然后再将<question, context>作为模型的输入，得到各知识点的 answer。

使用上面的方法可以较好地处理同一公司、同一类型的保险条款，这是因为同公司的保险条款文章结构具有一致性，可以用同一程序来分析上下文，但是对于不同公司、类型的保险条款，由于术语和结构不同，原分析程序无法处理，而为每种条款重新编写一份文本分析程序是不具可行性的，因此该模型需要改进。

### 2.5 模型改进

为了使知识抽取过程具有更好的泛用性，我们首先修改预测过程：将新条款的原文按照字数进行分段，每段约 300 字（尽量不打碎句子），而后将每一文本段都作为任何一个知识点的可能的 context，作为模型的输入。如果输出的 answer 是空，则说明这一段不存在相应的知识点，否则综合考虑每个知识点在所有文本段下的输出，选择其中概率最高的作为该知识点的 answer。这种新的预测方法对于任何条款具有泛用性，不再需要进行额外的文本解析。

我们用这种方法测试了几个不同公司的条款，结果表明其在旧模型上的效果并不好，准确率的下降很明显。原因在于：改进前在训练时，每个知识点的上下文都是根据文档结构精确定位的，没有很多的负样本，导致模型也只能够通过精确定位的上下文来进行预测。一旦文本组织结构、标题格式发生变化，原有的文本解析程序就无法精确定位问题上下文，产生了很多干扰数据，影响了模型的效果。

因此模型的训练过程需要修改，在 2.4 节的训练集基础上，我们添加分段文本数据，即对训练集中每个条款按照同样的方式分段，如果该段包含该知识点标注的答案，则作为新的样本，否则作为负样本（answer 为空）。在实际测试中，如果将这些新的样本全部加入到训练集中，产生的训练数据过多，且负样本数量远超正样本。

为了平衡此过程，我们进一步作如下改进：对于每个知识点问题，如果该条款本身不包含该知识点（因为知识点是针对所有保险条款统一定义的，因此对于某份特定的条款而言，未必所有知识点都包含在其中），则每个片段以 10%的概率作为该问题的负样本；如果条款本身包含该知识点，则分两种情况，如果当前文本片段包含目标知识点，则作为正样本，否则以 50%的概率选取为负样本，如此构造新的训练集得到新模型。这样做的想法是：如果条款中包含某一知识点，就增加与该知识点相关的负样本数量，以期模型能够更好地处理相似片段的干扰，提高答案的准确性。而如果条款中本身不包含该知识点，则文本片段与知识点的契合程度应该较差，选取少量的负样本就足够。

经过测试，新的模型相对于旧模型有很大改进，更契合新的预测方法，可以作为更通用的保险条款知识抽取模型。

## 3 实验

### 3.1 数据集

我们的数据集由某保险公司的保险条款组成，每个条款具有人工标注的知识点，如犹豫期，诉讼期，保险金额等。在实验过程中，训练集，测试集分别由 251 个条款和 98 个条款组成。经过统计，这些条款中所有可能的知识点问题数量为 309 条，平均每个条款有 45 条知识点需要提取。

### 3.2 评价指标

测试过程中，我们将条款文本分段，尝试从所有段中提取知识点$k_i$，并根据模型输出的概率，选择概率最高的文本作为该知识点的答案。如果最终得到的输出为空字符串，则代表条款不存在该知识点。由于每个条款提取的知识点只占 309 条中的小部分，大多数知识点的输出应当是空的，因此我们在评估时忽略这部分空知识点，关注两个指标：模型输出的知识点正确率$P$，即精准率（precision），以及应提取知识点中确实被正确提取的比率$R$，即召回率（recall）。假设知识点$k_i$标注为$y_i$，模型的输出为$\widetilde{y_i}$，则$P$和$R$可定义为：

$$P = \frac{\sum_i \mathbb{I}(\widetilde{y_i} \neq None, \widetilde{y_i} = y_i)}{\sum_i \mathbb{I}(\widetilde{y_i} \neq None)}$$

$$R=\frac{\sum_i \mathbb{I}(y_i \neq None, \tilde{y_i}=y_i)}{\sum_i \mathbb{I}(y_i \neq None)}$$

其中$\mathbb{I}(x)$定义为：

$$\mathbb{I}(x)=\begin{cases}1, & x=true\\0, & x=false\end{cases}$$

最后计算相应的$F_1$值，即准确率与召回率的调和平均值：

$$F_1=\frac{2PR}{P+R}$$

### 3.3 实验设置

我们使用 Google 开源的 BERT 中文预训练模型 BERT_chinese_L-12_H-768_A-12，在此基础上进行后续的测试。参数设置上，初始学习率为 3E-5，批量大小为 4，训练 epoch 数为 4，其余参数采用该模型的默认配置。

### 3.4 实验结果与分析

本文的实验包含两部分的测试，第一部分是基准模型的测试，其训练过程为：按 2.4 节的描述，首先使用文本解析程序解析保险条款的结构，提取出对应知识点所在的 context 后，再组合成训练集对 BERT 模型作 fine-tuning。第二部分是新模型的测试，其训练过程为：在基准模型的训练集的基础上添加新样本。对相应的保险条款按字数进行分段，每段文本约 300 字。对于每个知识点问题，按照 2.5 节的描述构建训练集来训练得到新模型。最后以 2.5 节描述的方式进行测试，测试结果是测试集中 98 个保险条款统计的平均值，如下表所示：

**表 1 实验结果**

| 模型 | $P$ | $R$ | $F_1$ |
|---|---|---|---|
| 新模型 | 63.37% | 54.61% | 58.66% |
| 基准模型 | 23.70% | 31.20% | 26.94% |

可以看出，以前文所述的方法添加有限的负样本后训练的模型明显优于基准模型，其中$P$提高了约 40%，$R$提高了约 20%。$P$的提升相当显著。如 2.5 节所述：基准模型的训练集中，仅通过文本解析程序精确定位知识点的上下文信息，导致模型只具备从正确的上下文中抽取对应的知识点的能力，而不具备辨别无效上下文的能力，因此基准模型存在很大比例的无效输出。而按比例添加负样本后，新模型的无效输出大幅度减少，输出的知识点中 60%以上是有效且正确的输出。而由于添加了相对于基准模型粒度更粗的上下文信息（文本段）组成的正样本，使得模型能够更好地从无规则截取的文本段中抽取出目标知识点，因此召回率$R$也有大幅提升。最终$F_1$值提升了约 30%。

### 3.5 总结与改进

实验结果表明，经过我们优化训练集后训练得到的新模型在文本分段预测的方法中效果优于原始的基准模型，从而能够进一步用于更具泛用性的保险条款知识抽取任务中。同时，当前的模型依然还有很大的提升空间。

（1）由于现实条件（数据标注量）的限制，我们的训练仅囊括了 251 个条款，且所有训练数据均来自于同一个保险人。在扩大数据集规模，囊括更多保险人制定的条款数据后，应当进一步优化模型的效果。

（2）当前我们的数据标注仅包含条款知识点的内容，而训练数据中对应的上下文是通过自主编写的文本分析程序得到的，这样得到的上下文存在小部分错误。可优化人工标注策略，同时标注知识点及其上下文，这样得到的数据可以更精确。

## 4 结论和未来的工作

本文提出了一种采用语言理解模型从非结构化领域文本中抽取知识的方法，旨在解决从特定领域专业文本中自动抽取特定信息的问题。所采用的语言理解模型基于 BERT，针对少量人工标注的数据进行了微调。根据实验显示，微调后的模型知识抽取的正确率过半，在工程项目中达到了基本可用的程度。

鉴于当前的模型是以一位保险人所提供的，通用性不强。下一步我们将引入更多保险人的同类型文档，并如 3.5 所述改进标注方法，按照本文所述方案训练出更具可扩展性且性能更优的模型。

## 参考文献

**作者：**

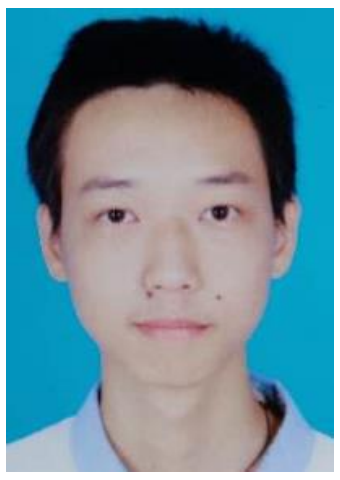

王梓嘉（1997—），硕士研究生，主要研究领域为自然语言处理、知识图谱。
E-mail：zijiaw@protonmail.com

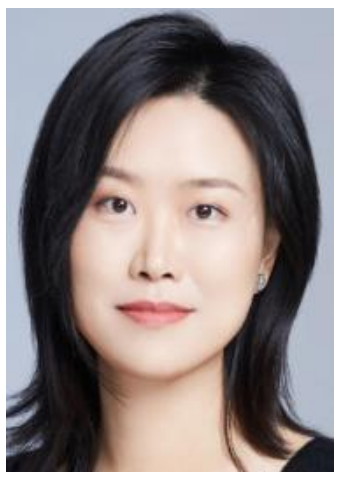

李烨（1977—），通信作者,硕士，资深算法工程师，主要研究领域为知识图谱、自然语言处理。

E-mail：jull@microsoft.com

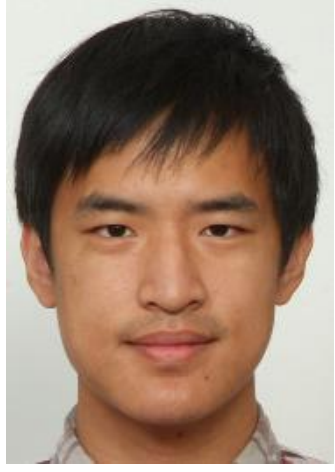

朱忠凯（1984—），硕士，资深算法工程师，主要研究领域为知识图谱、智能对话系统。

E-mail：zhonz@microsoft.com